\documentclass{article}
\makeatletter
\IfFileExists{arxiv/submission/macros.tex}
  {\def\subpath{arxiv/}\def\input@path{{arxiv/}}}
  {\def\subpath{}}
\makeatother
\usepackage{iclr2027_conference,times}
\usepackage{amsmath,amssymb,booktabs,tabularx,array,multirow,makecell}
\usepackage{graphicx,xcolor,microtype,xspace,wrapfig,enumitem,longtable}
\usepackage[most]{tcolorbox}
\usepackage{titletoc}
\usepackage{placeins}
\usepackage{tikz}
\usetikzlibrary{arrows.meta,positioning,calc,fit,backgrounds,shapes.misc}
\usepackage{hyperref,url}
\hypersetup{colorlinks=true,citecolor=blue,linkcolor=black,urlcolor=blue,
  pdftitle={Timeline-Bench: Evaluating Agents on Realistic Video-Editing Tasks, from Raw Footage to Final Cut},
  pdfauthor={Gunin Gupta, Nirmit Arora, Pavan Kalyan Tankala},
  pdfsubject={Benchmark for AI agents on end-to-end video editing},
  pdfkeywords={video editing, AI agents, benchmark, multimodal evaluation, human preference},
  pdfcreator={},pdfproducer={}}
\pdftrailerid{}
\graphicspath{{\subpath submission/figures/}}

\newcommand{\benchmark}{\textsc{Timeline-Bench}\xspace}
\newcommand{\xhdr}[1]{\smallskip\noindent{\bf #1.}}
\newcommand{\code}[1]{\texttt{\small #1}}
\newcommand{\task}[1]{\textbf{#1}}

\newcolumntype{Y}{>{\raggedright\arraybackslash}X}
\newcolumntype{R}{>{\raggedleft\arraybackslash}X}

\newcommand{\appref}[1]{Appendix~\ref{#1}}
\newcommand{\figpanel}[2]{\hyperref[#1]{Figure~\ref*{#1}#2}}
\definecolor{ink}{HTML}{213547}
\definecolor{tbgray}{HTML}{F4F5F7}
\definecolor{tbedge}{HTML}{8A94A6}
\tcbset{tbbox/.style={colback=tbgray,colframe=tbedge,boxrule=0.4pt,arc=1pt,left=4pt,right=4pt,top=3pt,bottom=3pt,fontupper=\small}}
\setlist{nosep,leftmargin=*}

\makeatletter
\def\@maketitle{\vbox{\hsize\textwidth
{\LARGE\sc \@title\par}
    \def\And{\end{tabular}\hfil\linebreak[0]\hfil
            \begin{tabular}[t]{l}\bf\rule{\z@}{24pt}\ignorespaces}%
    \def\AND{\end{tabular}\hfil\linebreak[4]\hfil
            \begin{tabular}[t]{l}\bf\rule{\z@}{24pt}\ignorespaces}%
    \begin{tabular}[t]{l}\bf\rule{\z@}{24pt}\@author\end{tabular}%
\vskip 0.3in minus 0.1in}}
\def\iclrruler#1{}%
\makeatother

\title{\benchmark: Evaluating Agents on\\Realistic Video-Editing Tasks,\\from Raw Footage to Final Cut}
\author{Gunin Gupta \qquad Nirmit Arora \qquad Pavan Kalyan Tankala \\
\normalfont TensorTest (Ritivel Labs Inc.) \\
\normalfont\texttt{founders@ritivel.com}}

\begin{document}
\maketitle
\begin{abstract}
AI agents increasingly carry out long-horizon professional work, but their evaluations rarely require a finished creative deliverable.
To this end, we introduce \benchmark, a benchmark of 56 real video-editing tasks, each asking an agent to turn raw production material into a finished video.
Tasks range from selecting dialog takes and shaping interview footage into a story to cutting commercials from product shots, voiceovers and graphics.
Every task provides a brief, source assets, a container and a set of tests.
A task is resolved when the output passes every test.
The tests check the delivery format, the content and the brief's explicit requirements, and include a quality test calibrated on 2,582 blind judgments by 43 video editors.
We evaluate 16 agents that pair frontier models with coding-agent harnesses such as Codex, Claude Code and OpenCode.
The best, GPT-6 Astra in Codex with curated editorial guidance, resolves only 15 of the 56 tasks (26.8\%), and the average agent resolves 14.0\%.
Human editors prefer the reference edit in 83.5\% of judgments.
Most unresolved runs (562 of 771) fail only the quality test: agents perceive footage through stills and transcripts and check their renders for defects, not craft.
We release the tasks, verifier and per-run results at \url{https://timelinebench.tensortest.com}.
\end{abstract}

\section{Introduction}
\label{sec:introduction}
Recent frontier models have substantially advanced the ability of AI agents to perform software engineering, scientific computing, and professional knowledge work \citep{openai2026astra,anthropic2026fable}. Models such as GPT-6 Astra and Claude Fable 5.1 combine stronger reasoning with improved computer use and sustained problem solving, enabling agents to carry out complex workflows across software applications. Systems such as Codex and Claude Code provide the tools to execute code, inspect intermediate results, and iteratively refine their work \citep{openai2025codex,anthropic2026claudecode}. These capabilities also extend to creative applications: for example, OpenAI demonstrates Astra modeling a house in Blender and turning it into an interactive scene in Unreal Engine \citep{openai2026astra}. As agents take on a wider range of professional work, benchmarks must assess their ability to complete realistic workflows and produce useful deliverables. Evaluations such as GDPval, Agents' Last Exam, and AutomationBench reflect this growing emphasis \citep{patwardhan2026gdpval,sun2026ale,shepard2026automationbench}.

Video editing presents a demanding environment for such an evaluation. Used in filmmaking, advertising, education, and digital media, it requires interpreting a brief, understanding source footage, and coordinating picture, speech, music, and graphics over time. These decisions are interdependent: changing a shot can alter the meaning of the accompanying narration, while changing its duration can affect pacing and synchronization. Although broad professional benchmarks include media-related tasks \citep{patwardhan2026gdpval,sun2026ale}, their aggregate results provide limited insight into these editorial capabilities. A focused evaluation is therefore needed to assess whether agents can turn raw audiovisual material and a brief into a coherent finished video.

In this paper, we introduce \benchmark, a benchmark for evaluating agents on complete video-editing assignments, from raw production material to finished video. \benchmark assesses whether agents can carry out workflows encountered in filmmaking, documentary production, advertising, and digital media, including constructing a narrative from interviews, assembling scenes from multiple takes, and producing promotional videos from mixed audiovisual assets. Each task in the benchmark is defined by a collection of source material, a project-specific brief, and a reproducible execution environment, with a reference video reserved for evaluation. Agents must interpret the brief, inspect the available assets, select and arrange footage, coordinate picture and sound, and render the final deliverable. We score each output with tests of delivery, content and brief compliance, including a quality test calibrated on blinded human preference, and report the human judgments separately. This protocol accommodates multiple valid editorial solutions while assessing both explicit requirements and viewing experience. Together, these assignments test audiovisual understanding, temporal reasoning, sustained tool use, and the interdependent creative decisions required to turn raw material into a coherent finished edit.

\begin{figure}[t]
  \centering
  \begin{minipage}[t]{0.49\linewidth}
  \vspace{0pt}
  \includegraphics[width=\linewidth]{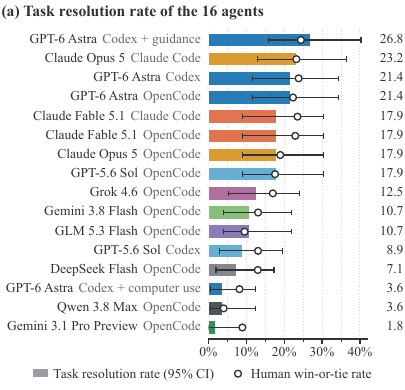}
  \end{minipage}\hfill
  \begin{minipage}[t]{0.49\linewidth}
  \vspace{0pt}
  \centering
\begin{tikzpicture}[x=0.172cm,y=0.172cm,font=\scriptsize]
\foreach \v in {10,20} {\draw[black!13,line width=0.3pt] (\v,0) -- (\v,30); \draw[black!13,line width=0.3pt] (0,\v) -- (30,\v);}
\draw[black!75,line width=0.4pt] (0,30) -- (0,0) -- (30,0);
\foreach \v in {0,10,20,30} {\draw[black!75,line width=0.4pt] (\v,0) -- (\v,-0.5); \node[anchor=north,inner sep=1pt] at (\v,-0.6) {\v}; \draw[black!75,line width=0.4pt] (0,\v) -- (-0.5,\v); \node[anchor=east,inner sep=1pt] at (-0.6,\v) {\v};}
\draw[black!55,dashed,line width=0.5pt] (0,0) -- (30,30);
\node[anchor=north] at (15,-2.6) {Task resolution rate (\%)};
\node[anchor=south,rotate=90] at (-3.1,15) {Human win-or-tie rate (\%)};
\filldraw[shift={(21.43,22.33)},fill={rgb,255:red,39;green,123;blue,183},draw=black!70,line width=0.3pt] (-1.45pt,-1.45pt) rectangle (1.45pt,1.45pt); %
\filldraw[shift={(17.86,22.92)},fill={rgb,255:red,226;green,116;blue,79},draw=black!70,line width=0.3pt] (-1.45pt,-1.45pt) rectangle (1.45pt,1.45pt); %
\filldraw[shift={(17.86,18.95)},fill={rgb,255:red,216;green,156;blue,50},draw=black!70,line width=0.3pt] (-1.45pt,-1.45pt) rectangle (1.45pt,1.45pt); %
\filldraw[shift={(17.86,17.58)},fill={rgb,255:red,101;green,176;blue,211},draw=black!70,line width=0.3pt] (-1.45pt,-1.45pt) rectangle (1.45pt,1.45pt); %
\filldraw[shift={(12.50,16.99)},fill={rgb,255:red,178;green,110;blue,159},draw=black!70,line width=0.3pt] (-1.45pt,-1.45pt) rectangle (1.45pt,1.45pt); %
\filldraw[shift={(10.71,13.10)},fill={rgb,255:red,135;green,190;blue,121},draw=black!70,line width=0.3pt] (-1.45pt,-1.45pt) rectangle (1.45pt,1.45pt); %
\filldraw[shift={(10.71,9.52)},fill={rgb,255:red,138;green,123;blue,185},draw=black!70,line width=0.3pt] (-1.45pt,-1.45pt) rectangle (1.45pt,1.45pt); %
\filldraw[shift={(7.14,13.03)},fill={rgb,255:red,132;green,135;blue,141},draw=black!70,line width=0.3pt] (-1.45pt,-1.45pt) rectangle (1.45pt,1.45pt); %
\filldraw[shift={(3.57,4.00)},fill={rgb,255:red,79;green,85;blue,91},draw=black!70,line width=0.3pt] (-1.45pt,-1.45pt) rectangle (1.45pt,1.45pt); %
\filldraw[shift={(1.79,8.95)},fill={rgb,255:red,48;green,147;blue,87},draw=black!70,line width=0.3pt] (-1.45pt,-1.45pt) rectangle (1.45pt,1.45pt); %
\filldraw[shift={(23.21,23.21)},fill={rgb,255:red,216;green,156;blue,50},draw=black!70,line width=0.3pt] (90:2.1pt) -- (210:2.1pt) -- (330:2.1pt) -- cycle; %
\filldraw[shift={(21.43,23.81)},fill={rgb,255:red,39;green,123;blue,183},draw=black!70,line width=0.3pt] (0:2.0pt) -- (60:2.0pt) -- (120:2.0pt) -- (180:2.0pt) -- (240:2.0pt) -- (300:2.0pt) -- cycle; %
\filldraw[shift={(17.86,23.51)},fill={rgb,255:red,226;green,116;blue,79},draw=black!70,line width=0.3pt] (90:2.1pt) -- (210:2.1pt) -- (330:2.1pt) -- cycle; %
\filldraw[shift={(8.93,13.10)},fill={rgb,255:red,101;green,176;blue,211},draw=black!70,line width=0.3pt] (0:2.0pt) -- (60:2.0pt) -- (120:2.0pt) -- (180:2.0pt) -- (240:2.0pt) -- (300:2.0pt) -- cycle; %
\filldraw[shift={(26.79,24.40)},fill={rgb,255:red,39;green,123;blue,183},draw=black!70,line width=0.3pt] (0:2.0pt) -- (90:2.0pt) -- (180:2.0pt) -- (270:2.0pt) -- cycle; %
\filldraw[shift={(3.57,8.18)},fill={rgb,255:red,39;green,123;blue,183},draw=black!70,line width=0.3pt] (90:2.3pt) -- (126:0.95pt) -- (162:2.3pt) -- (198:0.95pt) -- (234:2.3pt) -- (270:0.95pt) -- (306:2.3pt) -- (342:0.95pt) -- (18:2.3pt) -- (54:0.95pt) -- cycle; %
\node[anchor=south east,inner sep=1pt] at (28.39,25.40) {Astra, curated guidance};
\draw[black!45,line width=0.3pt] (1.79,9.50) -- (1.79,13.85);
\node[anchor=south west,inner sep=1pt,align=left] at (0.79,13.95) {Gemini 3.1\\Pro Preview};
\node[anchor=west,inner sep=0.6pt,fill=white] at (4.47,7.48) {Computer use};
\node[anchor=west,inner sep=0.6pt,fill=white] at (4.47,4.00) {Qwen 3.8 Max};
\filldraw[shift={(17.6,10.6)},fill=white,draw=black!70,line width=0.3pt] (-1.45pt,-1.45pt) rectangle (1.45pt,1.45pt); \node[anchor=west,inner sep=1pt] at (18.4,10.6) {OpenCode};
\filldraw[shift={(17.6,8.7)},fill=white,draw=black!70,line width=0.3pt] (0:2.0pt) -- (60:2.0pt) -- (120:2.0pt) -- (180:2.0pt) -- (240:2.0pt) -- (300:2.0pt) -- cycle; \node[anchor=west,inner sep=1pt] at (18.4,8.7) {Codex CLI};
\filldraw[shift={(17.6,6.8)},fill=white,draw=black!70,line width=0.3pt] (90:2.1pt) -- (210:2.1pt) -- (330:2.1pt) -- cycle; \node[anchor=west,inner sep=1pt] at (18.4,6.8) {Claude Code};
\filldraw[shift={(17.6,4.9)},fill=white,draw=black!70,line width=0.3pt] (0:2.0pt) -- (90:2.0pt) -- (180:2.0pt) -- (270:2.0pt) -- cycle; \node[anchor=west,inner sep=1pt] at (18.4,4.9) {Curated guidance};
\filldraw[shift={(17.6,3.0)},fill=white,draw=black!70,line width=0.3pt] (90:2.3pt) -- (126:0.95pt) -- (162:2.3pt) -- (198:0.95pt) -- (234:2.3pt) -- (270:0.95pt) -- (306:2.3pt) -- (342:0.95pt) -- (18:2.3pt) -- (54:0.95pt) -- cycle; \node[anchor=west,inner sep=1pt] at (18.4,3.0) {Computer use};
\node[anchor=south west,inner sep=0pt,font=\fontsize{7.5}{9}\selectfont\bfseries] at ([yshift=1pt]current bounding box.north west) {(b) Resolution rate against human win-or-tie rate};
\end{tikzpicture}
  \end{minipage}
  \caption{\textbf{Even the best agent resolves about a quarter of the tasks.} \textbf{(a)} Task resolution rate of the 16 agents on \benchmark, one run per task, with exact 95\% confidence intervals; open circles give the human win-or-tie rate (\autoref{sec:human-study}). \textbf{(b)} The same two rates per agent, with quality margins held out by agent (\autoref{sec:results}). Colors follow (a), marker shapes give the harness or condition, and the dashed line marks equal rates.}
  \label{fig:leaderboard}
  \label{fig:auto-vs-human}
\end{figure}
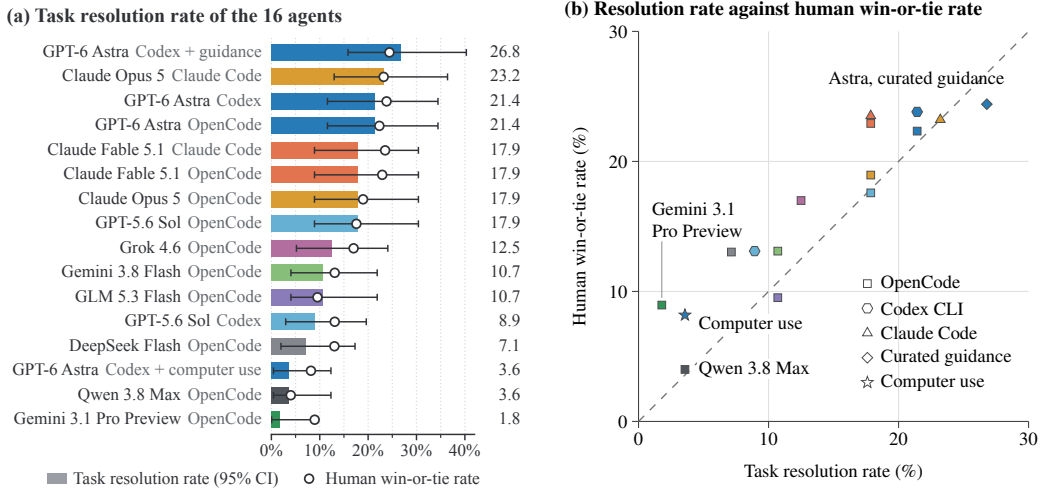

The remainder of this paper is structured as follows. We first describe \benchmark{}'s construction and evaluation protocol. We then benchmark frontier LLMs and agents on 56 editing assignments. The best agent resolves 15 of the 56 tasks (\figpanel{fig:leaderboard}{a}). Finally, we analyze failure modes to inform future LLM and agent development, and compare harnesses, editorial guidance and computer use in \appref{app:results}.

\section{\benchmark}
\label{sec:benchmark}
\label{sec:task-formulation}

\begin{figure}[t]
  \centering
  \includegraphics[width=\linewidth]{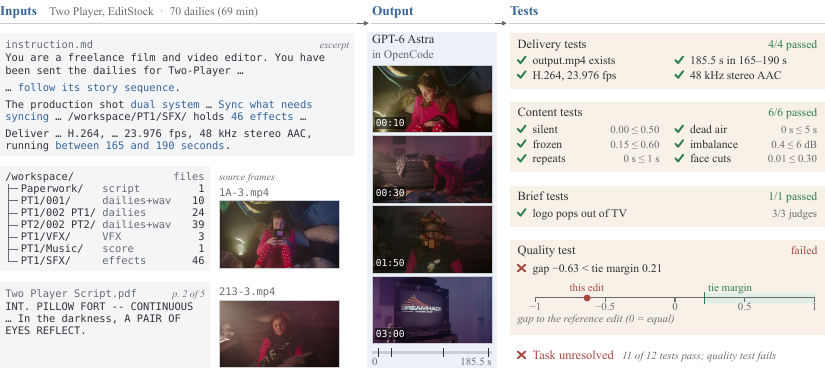}
  \caption{\textbf{An example \benchmark task, \task{Two Player} (EditStock).} Left: an excerpt of the brief the agent reads (\code{instruction.md}) and its production workspace. Middle: frames from the 185-second edit that GPT-6 Astra produced in OpenCode. Right: the tests of this run.}
  \label{fig:task}
\end{figure}

A \benchmark task consists of an edit brief, source assets, a Docker image, a set of tests, and a time limit (\autoref{fig:task}). The brief describes the video the agent must produce, while the Docker image provides the tools and dependencies needed to complete the assignment. The tests assess whether the rendered video meets the technical specifications and satisfies the brief's content requirements. A task is successfully completed when all required tests pass and the benchmark performance is measured by the percentage of assigned tasks that are successfully completed. Evaluation focuses on the final video, allowing agents to choose their own workflow and produce different valid edits. Given the brief and source assets, an agent must inspect the material, select and arrange clips, coordinate the picture and sound, and export the requested video within the allotted time.

\subsection{Dataset Construction}
\label{sec:dataset-construction}

We construct \benchmark from 56 editing assignments drawn from three sources: 11 purchased project packages from EditStock\footnote{\url{https://editstock.com}}, 15 publicly available editing projects from Cinestudy\footnote{\url{https://cinestudy.org}}, and 30 newly commissioned projects from a professional video-editing agency. The agency supplied raw footage and finished reference edits. For Cinestudy, a professional editor selected reference edits from submissions linked publicly in the project-page comments, considering editing quality and availability. We prepare each assignment by organizing the available footage, audio, graphics, and supporting documents, and writing a brief that specifies the editorial objectives and delivery requirements. 

\subsection{Verification}
\label{sec:verification}

We consider a task verified when its assignment is internally consistent, each of its tests has been validated against outputs with a known reference, and the agent has no access to the reference edit.

\xhdr{Assignment review} A task is well-specified if its brief describes an edit that the source assets can support and that the reference edit exemplifies. To confirm this, each brief was revised over several review rounds, after which a professional editor checked all 56 assignments for consistency between the brief, the source assets, and the reference edit (\appref{app:qc}).

\xhdr{Test validation} A test is valid if it accepts outputs that have the property it checks and rejects outputs that do not. Each task therefore ships a mechanical oracle, a deliberately low-craft render of the source assets that meets the delivery specification, on which the delivery tests (\autoref{sec:resolution}) are validated. The content and brief tests are validated on the reference edit, which passes all the content tests and all the brief tests. To confirm that these tests also reject defective outputs, we construct single-defect controls from real edits, for example, by muting the sound, blacking out the opening, reordering sections, or freezing the picture.

\xhdr{Reference isolation} An agent should not be able to pass a task by recovering its reference edit. Reference edits are stored separately from the task inputs and are never staged in the container. Agents may use the internet to consult tool documentation, but the briefs prohibit retrieving finished edits, and the traces of all agent runs contain no web fetch and no access to a reference edit (\appref{app:qc}).

\subsection{Composition}
\label{sec:composition}

\benchmark tasks vary widely. For example, \task{Klug Brand Story} asks for a sixty-second, interview-driven brand-story commercial from about four hours of dailies. \task{Beauty of Delhi} asks for a heritage travel commercial from 603 photographs and 14.9 seconds of camera footage, built largely from still-image sequences, photo holds and controlled crop movement. \benchmark comprises approximately 33.0 hours of primary source footage; source duration ranges from 2.3 to 240.4 minutes (median 12.4), and the 41 landscape and 15 portrait deliverables have duration windows from 25 to 315 seconds (per-collection statistics in \appref{app:tasks}).

\begin{wrapfigure}[16]{r}{0.48\linewidth}
  \vspace{-5pt}
  \centering
  \includegraphics[width=\linewidth]{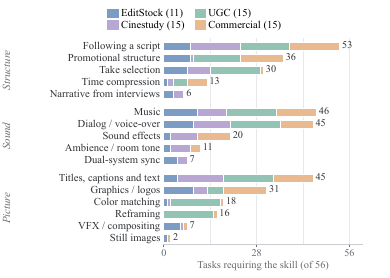}
  \vspace{-16pt}
  \caption{\textbf{Editing skills required across tasks.} Each bar counts the tasks whose brief, paperwork or supplied material requires the skill.}
  \label{fig:skills}
  \vspace{-10pt}
\end{wrapfigure}
The assignments span narrative scenes, documentaries, interviews, action sequences, trailers, product advertisements, personal-branding videos, and lifestyle and travel films. A typical task requires seven of the 16 editing skills in \autoref{fig:skills} (range 1--10), and 54 of the 56 tasks need at least one skill from each family. Following a script (53 tasks), music (46), dialog or voice-over editing (45) and titles, captions, or other on-screen text (45) are required almost everywhere, whereas take selection (30), sound effects (20), dual-system sync (7) and VFX compositing (7) appear in fewer tasks. The collections differ: every UGC task requires captions, color matching and reframing for portrait delivery, which almost no other task does, whereas Cinestudy and EditStock tasks hold most of the ambience, dual-system sync and VFX work.

\section{Evaluation}
\label{sec:evaluation}

Agent benchmarks often decide success with tests alone, because their tasks specify an acceptable end state \citep{jimenez2024swebench,merrill2026terminalbench}. In video editing, many different edits satisfy the same brief, so tests of its requirements are necessary but cannot establish quality. We therefore evaluate each output in three ways. First, tests check that the edit is a valid delivery and meets the brief's explicit requirements (\autoref{sec:resolution}). Second, a blind study of human preference has video editors compare each output with its task's reference edit (\autoref{sec:human-study}). Third, a quality test has multimodal LLM judges make the same comparison, calibrated on the human judgments (\autoref{sec:quality-test}); an output that passes it and all other tests resolves its task.

\subsection{Task Resolution Rate}
\label{sec:resolution}

An output \emph{resolves} its task only if it passes every test. The tests inspect only the delivered video and are of four kinds. Four \emph{delivery tests} per task check that the file exists and that its video stream, duration and audio stream match the brief. Six \emph{content tests}, shared by all tasks, reject degenerate edits such as mostly silent, frozen or looped ones (\appref{app:tests}). \emph{Brief tests}, one to six per task and 180 in total, check what the brief explicitly requires, such as using a supplied voice-over or keeping a scene order. Of these, 153 are programmatic: code matches the edit's frames and soundtrack against the task's source clips and recordings to find which ones it uses and where, reads its on-screen text with OCR (PP-OCR via RapidOCR; \citealp{du2020ppocr}) and transcribes its speech (Whisper \texttt{small} via faster-whisper; \citealp{radford2023whisper}). Each such test then checks one of these facts, for example that a required line is spoken, that the supplied voice-over is heard or that clips appear in the required order. The other 27 brief tests concern visible content and are decided by a two-of-three majority of model judges (Gemini 3.8 Flash, GPT-6 Astra and Claude Opus 5.5; \citealp{google2026gemini38flash,openai2026astra,anthropic2026opus55}; \appref{app:brief}). Finally, the \emph{quality test} compares the edit with its task's reference edit (\autoref{sec:quality-test}). Each agent runs once on each of the 56 tasks, and its \emph{task resolution rate} is the share of tasks it resolves. A missing output resolves nothing.

\subsection{Human Preference Study}
\label{sec:human-study}

In a blind pairwise study, 43 professional video editors compared each delivered output with its task's reference edit. They were paid hourly, independently of their answers, and report an average of at least two years of professional experience (\appref{app:human}). The two edits appear as ``Version A'' and ``Version B'' in random order (\autoref{fig:study-compare}), next to a one-line statement of the project's audience and purpose. Human editors do not see the brief, the task title, the agent or other human editors' answers. They answer ``Considering the complete viewing experience, which version would you choose for this project?'' with \emph{A preferred}, \emph{B preferred} or \emph{no meaningful preference}, or mark the pair \emph{cannot assess} with a reason; we exclude those. Answers unlock only after both videos have played to the end at normal speed with sound, and the seek bar stays hidden until each video's first full viewing. Three distinct human editors judge each of the 863 pairs, giving 2,589 judgments, of which 2,582 are marked assessable. An agent's \emph{win-or-tie rate}, as in GDPval \citep{patwardhan2026gdpval}, is the share of judgments in which a human editor prefers its edit or has no meaningful preference, averaged within each task and then over tasks.

\subsection{Quality Test}
\label{sec:quality-test}

The quality test asks whether an edit is at least as good as its task's reference edit. Three multimodal LLM judges, Gemini 3.8 Flash, GPT-6 Astra and Claude Opus 5.5, score each edit separately, without seeing the other version. Like the human editors, they see the one-line statement of audience and purpose but not the brief. Additionally, they receive the study's written rubric of what makes an edit acceptable (\appref{app:quality-judges}) and rate the edit from 1 to 10 overall and on story and assembly, pacing, picture, sound and graphics, which we combine into one score (\appref{app:quality-scoring}). Gemini 3.8 Flash watches the edit with sound. GPT-6 Astra and Claude Opus 5.5 accept only text and images, so they read contact sheets of one frame per second instead, together with four measurements computed from the file: cuts per minute from a shot-boundary detector (our re-implementation of PySceneDetect's adaptive detector; \citealp{castellano2026pyscenedetect}), the longest stretch of static picture, integrated loudness (EBU R128, measured with FFmpeg) and the share of silent runtime. Because judges use the 1-to-10 scale differently, we convert each judge's scores to z-scores, using that judge's mean and standard deviation over the 919 videos in the study (863 agent edits and 56 reference edits), and keep these constants fixed for new submissions. The \emph{panel score} of an edit is the average of its three z-scores.

The test passes when the agent edit's panel score exceeds the reference edit's by at least the \emph{tie margin} of the task's collection. A margin is fitted on a collection's study pairs: it is set so that the test's pass rate comes as close as possible to the human win-or-tie rate without exceeding it. Because the panel score averages z-scores, margins are measured in judge standard deviations; one judge standard deviation is 1.0 to 1.5 points on the 1-to-10 scale.  We fit margins in two ways. For new submissions, which have no human judgments, the margins are fitted once on all 2,582 assessable judgments and then frozen: 0.30 (Cinestudy), 0.52 (Commercial), 0.21 (EditStock) and 1.54 (UGC) (\appref{app:quality}). For the results we report, fitting on an agent's own judgments would be circular, so each agent is scored with margins refitted per collection without its own judgments (leave-one-agent-out). These held-out margins stay within 0.09 of the frozen values. All margins are positive, so an agent edit must score higher than its reference edit to pass.

\section{Experimental Setup}
\label{sec:setup}
We evaluate 16 agents. An \emph{agent} is a model running in a \emph{harness}, the program that gives the model its tools and executes its commands. Each agent attempts each task once, giving 896 \emph{runs}.

\subsection{Agents}
\label{sec:agents}
An agent's result depends on both its model and its harness, and model developers tune their own harnesses for their own models. We therefore run ten models in one open-source harness, OpenCode 1.18.31~\citep{opencode2026}: GPT-6 Astra~\citep{openai2026astra}, GPT-5.6 Sol~\citep{openai2026gpt56}, Claude Fable 5.1~\citep{anthropic2026fable}, Claude Opus 5~\citep{anthropic2026opus5}, Gemini 3.1 Pro Preview~\citep{google2026gemini31pro}, Gemini 3.8 Flash~\citep{google2026gemini38flash}, DeepSeek Flash~\citep{deepseek2026v41flash}, Grok 4.6~\citep{xai2026grok46}, GLM 5.3 Flash~\citep{zai2026glm53flash} and Qwen 3.8 Max~\citep{qwen2026qwen38max}. Four of these models also run in their developers' own harnesses: GPT-5.6 Sol and GPT-6 Astra in Codex CLI 0.155.1~\citep{openai2025codex}, and Claude Fable 5.1 and Claude Opus 5 in Claude Code 2.1.278~\citep{anthropic2026claudecode}. We use unmodified releases of all three harnesses, and every model runs at the highest reasoning effort its provider offers (\appref{app:runtime}).

\xhdr{Curated guidance} One agent adds editorial guidance to GPT-6 Astra in Codex CLI. It receives 1,909 words of general editing advice: a skill file (\code{SKILL.md}) and six reference notes on reading the brief and footage, planning and building the edit, reviewing and delivering it, and using the installed video tools. The advice contains no task-specific answers. A one-line note in the agent's instruction file (\code{AGENTS.md}) says where to find it; everything else matches the unguided agent.

\xhdr{Computer use} Another agent runs GPT-6 Astra in Codex CLI's computer-use mode, in which it sees the screen and controls the mouse and keyboard but has no shell, so it cannot use command-line tools. It edits in the free edition of DaVinci Resolve~\citep{blackmagic2026resolve} on macOS. Its briefs are the same except for the tool instructions, which tell it to work in Resolve, and a Resolve project is already open with the task's media imported.

\subsection{Environment and Budgets}
\label{sec:environment}
The coding agents run in Linux containers with the tools an editor working in code needs: FFmpeg, Python and Node, the programmatic video frameworks Remotion and HyperFrames with a headless browser, Poppler for PDF paperwork, and a transcription command (AssemblyAI Universal-3.5 Pro; versions in \appref{app:runtime}). Every run starts from a fresh harness state and the same input files, which are checked against their SHA-256 hashes before the run. Each run has a 300-minute limit and a whole machine to itself (32 vCPUs and 256\,GB of memory), with no CPU or memory limits on the container. Agents may use the internet, but the briefs forbid retrieving finished edits, and we audit the traces for such retrievals (\autoref{sec:verification}). Tasks are packaged in the Harbor format~\citep{harbor2026}, which fixes each task's brief, inputs and delivery tests
(\appref{app:runtime}).

\section{Results}
\label{sec:results}

\figpanel{fig:leaderboard}{a} shows the task resolution rate of every agent, and \autoref{tab:results} gives the numbers (with confidence intervals in \autoref{tab:detail-results}). GPT-6 Astra in Codex CLI with curated editorial guidance resolves the most tasks, 15 of 56 (26.8\%), followed by Claude Opus 5 in Claude Code at 23.2\% and by GPT-6 Astra in Codex CLI and in OpenCode at 21.4\% each. With one run per task, each agent's rate has a 95\% CI of about $\pm$11 points, so the agents near the top cannot be told apart. Overall, 125 of the 896 runs resolve their task (14.0\%, 95\% CI 11.7--16.4\%), and \autoref{sec:analysis} examines where the others fail. The median runtime per run ranges from 16 to 61 minutes and is unrelated to resolution (Spearman $\rho = 0.15$). Among the 12 agents with recorded costs, from \$0.13 to \$41.27 per run, more expensive agents tend to resolve more tasks ($\rho = 0.79$), but the best agent, GPT-6 Astra with curated guidance, costs an estimated \$9.76 per run (Codex CLI records only tokens; \appref{app:cost}).

\begin{table}[t]
\caption{\textbf{Main results}, one run per task. \emph{Tests}: runs, of 56, passing every test but the quality test. \emph{Res.}: resolution rate. \emph{Human}: human win-or-tie rate. With CIs in \autoref{tab:detail-results}.}
\label{tab:results}
\centering
\scriptsize
\setlength{\tabcolsep}{2pt}
\begin{tabular*}{0.487\linewidth}[t]{@{\extracolsep{\fill}}llrrr@{}}
\toprule
Model & Harness & Tests & Res.\,\% & Human\,\% \\
\midrule
GPT-6 Astra & OpenCode & 48 & 21.4 & 22.3 \\
Claude Fable 5.1 & OpenCode & 50 & 17.9 & 22.9 \\
Claude Opus 5 & OpenCode & 46 & 17.9 & 19.0 \\
GPT-5.6 Sol & OpenCode & 45 & 17.9 & 17.6 \\
Grok 4.6 & OpenCode & 39 & 12.5 & 17.0 \\
Gemini 3.8 Flash & OpenCode & 42 & 10.7 & 13.1 \\
GLM 5.3 Flash & OpenCode & 41 & 10.7 & 9.5 \\
DeepSeek Flash & OpenCode & 43 & 7.1 & 13.0 \\
Qwen 3.8 Max & OpenCode & 35 & 3.6 & 4.0 \\
Gemini 3.1 Pro Preview & OpenCode & 19 & 1.8 & 9.0 \\
\bottomrule
\end{tabular*}\hfill
\begin{tabular*}{0.487\linewidth}[t]{@{\extracolsep{\fill}}llrrr@{}}
\toprule
Model & Harness & Tests & Res.\,\% & Human\,\% \\
\midrule
Claude Opus 5 & Claude Code & 50 & 23.2 & 23.2 \\
GPT-6 Astra & Codex CLI & 48 & 21.4 & 23.8 \\
Claude Fable 5.1 & Claude Code & 49 & 17.9 & 23.5 \\
GPT-5.6 Sol & Codex CLI & 45 & 8.9 & 13.1 \\
\midrule
GPT-6 Astra & Codex CLI, guidance & 52 & 26.8 & 24.4 \\
\midrule
GPT-6 Astra & Codex CLI, comp.\ use & 35 & 3.6 & 8.2 \\
\midrule
All agents &  & 687 & 14.0 & 16.5 \\
\bottomrule
\end{tabular*}
\end{table}

\label{sec:human-results}
\label{sec:auto-vs-human}
\label{sec:cost}
\xhdr{Human preference}
Human editors prefer the reference edit in 83.5\% of the 2,582 assessable judgments, the agent edit in 11.1\%, and have no meaningful preference in 5.5\%. The win-or-tie rate averages 16.5\% over agents. It ranges from 24.4\% for GPT-6 Astra with curated guidance, 8.3 points of which come from ties, to 4.0\% for Qwen 3.8 Max.

Single judgments are noisy (Krippendorff's $\alpha = 0.10$), but per-agent rates are reliable (Spearman--Brown reliability 0.84 with three human editors per pair), robust to leaving out any human editor (Spearman $\rho \ge 0.96$) and unaffected by the order of the two versions (Appendices~\ref{app:human} and~\ref{app:reliability}). The resolution rate tracks the human win-or-tie rate across agents (\figpanel{fig:auto-vs-human}{b}). With margins refit without each agent's own judgments, the two rates differ by 3.0 points on average and by 7.2 at most (Gemini 3.1 Pro Preview), with Spearman $\rho = 0.93$ and Pearson $r = 0.93$ over the 16 agents.

\section{Analysis}
\label{sec:analysis}

In this section, we ask where runs fail, why human editors prefer the reference edits and how agents work (full analysis in \appref{app:ext}, methods in \appref{app:analysis}).

\begin{figure}[t]
\centering
\includegraphics[width=\linewidth]{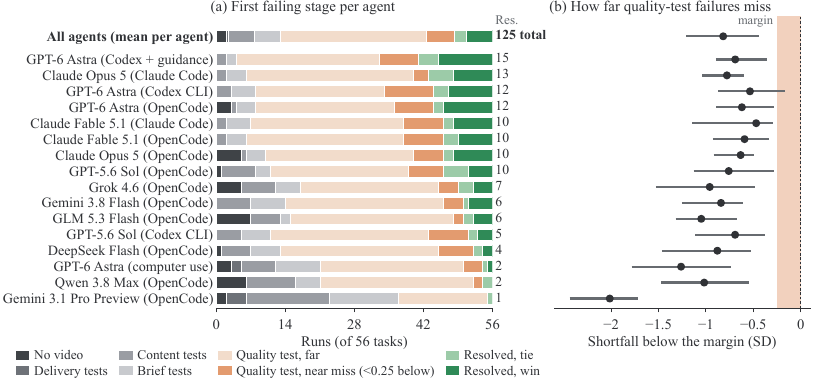}
\caption{\textbf{Most unresolved runs pass every test and fail only the quality test.} (a) The 56 runs of each agent by the first test they fail (\autoref{sec:resolution}); quality-test failures are split into near misses (less than 0.25 panel standard deviations below the margin) and far misses, resolved runs into wins and ties. (b) How far each agent's quality-test failures fall below the margin, in judge standard deviations: median (dot) and interquartile range (bar); the shaded band marks near misses.}
\label{fig:failure-stages}
\end{figure}

\xhdr{The quality test is where agents lose}
\label{sec:failure-funnel}
Of the 771 unresolved runs, 562 (73\%) pass the delivery, content and brief tests and fail only the quality test (\autoref{fig:failure-stages}a); for 15 of the 16 agents this accounts for 61--90\% of their losses, the exception being Gemini 3.1 Pro Preview, which loses 35 of its 55 unresolved runs to earlier tests. Most of these failures are narrow (\autoref{fig:failure-stages}b): the median one falls 0.82 judge standard deviations short of the margin, about one point on the 1-to-10 scale, and 0.47 to 0.78 for the eight agents that resolve at least 10 tasks, against 2.02 for Gemini 3.1 Pro Preview. Because so many runs sit near the bar, the number resolved depends on where the margin is set, but the order of agents largely does not: lowering every margin by 0.1 standard deviations would resolve 27 more runs, and although agents a run or two apart can swap places, including the top two, the ranking barely moves (Spearman 0.96).

\begin{figure}[t]
\centering
\includegraphics[width=\linewidth]{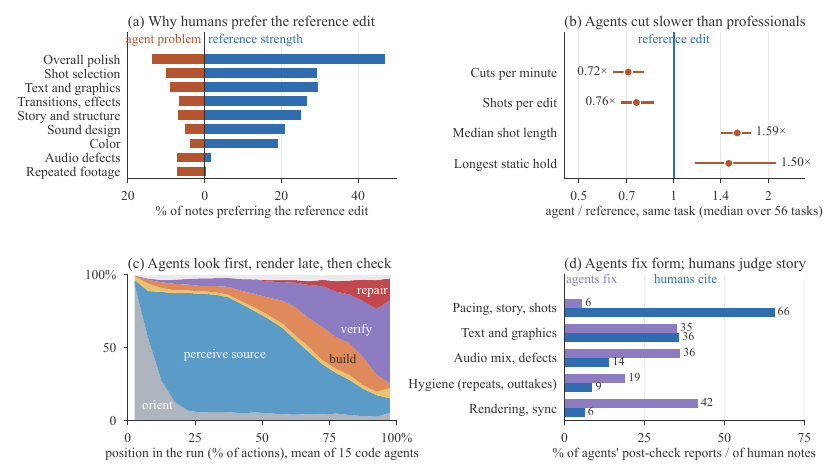}
\caption{\textbf{Agents produce clean assemblies but not finished edits, and check for defects rather than quality.} (a) Share of the human notes preferring the reference edit that name a problem of the agent edit or a strength of the reference edit. (b) Agent edit relative to the reference edit of the same task (median over tasks of the mean ratio; 95\% task-bootstrap interval). (c) Phases of the native actions of the 15 code agents over the course of a run. (d) What agents report fixing after checking their own render (343 reports) and what human editors cite in notes preferring the reference edit.}
\label{fig:analysis-compact}
\end{figure}

\xhdr{Human editors prefer the reference edit for craft that the agent edit lacks}
\label{sec:editors-say}
\label{sec:craft}
We coded all 1,271 human editors' notes with Claude Opus 5.5 into 24 categories (second-coder $\kappa = 0.84$, author-audited; \appref{app:editor-notes}). Of the 1,101 notes explaining a preference for the reference edit, 95\% cite a strength of the reference edit but only 44\% name any problem in the agent edit (\autoref{fig:analysis-compact}a). The strengths are finishing and editorial judgment, such as polish, shot selection, graphics, and transitions; only audio defects and repeated footage are main agent errors. Measurements agree: paired by task, agents change shots only 0.72 times as often as the reference edit and hold their longest static shot 1.5 times as long (\autoref{fig:analysis-compact}b), and human editors penalize the slower pace but not a faster one. Agents win on story, not on graphics (``[agent] gets into the heist story more directly''). Graphics decides 91\% of tagged UGC judgments, story 84--86\% in the narrative collections. Stronger agents fail more subtly. When humans reject an agent edit, their notes name a defect that a content test could catch, such as an audio glitch or repeated footage, for 10\% of the six highest-rated agents but for 19\% of the three weakest code agents (Gemini 3.1 Pro Preview, Qwen 3.8 Max and GLM 5.3 Flash). The computer-use agent is worst, at 28\%: it leaves slates, crew talk or retakes in 18.5\% of its rejected edits (``In [agent], you can hear the command `Action!' That's not okay.'').

\xhdr{Difficulty belongs to tasks}
\label{sec:difficulty}
Agents tend to fail the same tasks: 35 tasks are resolved by no agent, whereas only about 5 would be if each agent's successes fell on random tasks. Tasks also differ more than agents do. If we model each human vote as depending on the agent's skill and the task's difficulty (a Rasch model), task difficulty varies 2.6 times as much as agent skill, and even the best agent is more likely to lose than to win or tie on 54 of the 56 tasks (\appref{app:ext-difficulty}). The collection explains half of this: the human win-or-tie rate falls from 28.7\% on Cinestudy to 6.8\% on UGC. Within a collection, however, no task property or required editing skill predicts difficulty, and no agent is especially strong at particular skills.

\xhdr{Agents look first, render late and check for defects}
\label{sec:trajectory}
Code agents spend 54\% of their 111,636 native actions perceiving the source and 17\% verifying their own render, but only 9\% building; 98\% of their perception precedes the first render, which comes 72--90\% of the way through a run (\autoref{fig:analysis-compact}c). They see footage as stills (83\% of native reads open frames or contact sheets) and hear it as transcripts and level measurements; GPT agents in Codex CLI even print audio excerpts as base64 text, in up to 55 of 56 runs. Checking is nearly universal (86--100\% of runs for 14 of 15 code agents) and tracks resolution across agents ($\rho = 0.73$), but it targets form: only 5.5\% of the problems agents report after a check concern pacing, story or shot choice, against 66\% of human editors' notes (\autoref{fig:analysis-compact}d). Agents' final messages claim full success in 93\% of runs, including 95.5\% of edits that fail a test.

\begin{table}[t]
\caption{\textbf{Current agents as editors.} Evidence from \autoref{sec:analysis} and \appref{app:ext}.}
\label{tab:abilities}
\centering
\scriptsize
\setlength{\tabcolsep}{3pt}
\begin{tabularx}{\linewidth}{@{}lXl@{}}
\toprule
Ability & Evidence & Reading \\
\midrule
Delivery and explicit requirements & 857 of 863 videos meet the specification; 762 pass every brief test & Reliable \\
Sound & Required lines missed: 5.9\% (transcripts); 41.5\% for computer use & Via tools only \\
Timeline hygiene & Repeats, outtakes in 1--7\% of top agents' losses; 18.5\% for computer use & Mostly reliable \\
Checking one's own edit & Output checked in 86--100\% of runs; 5.5\% of fixes concern pacing, story, shots & Defects only \\
Judging one's own edit & 93\% of runs claim success, including 95.5\% of edits that fail a test & Absent \\
Story assembly & Drives agent wins; narrative scenes are the best collection (28.7\%) & Emerging \\
Rhythm and shot selection & 0.72$\times$ the reference cut rate; shot selection cited in 29\% of losses & Weak \\
Finishing & Graphics, transitions, sound, color: reference strengths in 19--47\% of notes & Weak \\
\bottomrule
\end{tabularx}
\end{table}

\autoref{tab:abilities} summarizes that agents reliably do what a brief can state (delivering the specified video, meeting explicit requirements, keeping the timeline clean), but they cannot judge their own edits and remain weak at what a human editor adds: rhythm, shot selection and finishing.

\section{Limitations}
\label{sec:limitations}
Each agent runs once per task, so our intervals omit run-to-run variance. The quality test was calibrated on the same human editors' votes that it is validated against, and it is valid per agent, not per edit (\appref{app:quality}). The reference edits are professional edits, not certified ground truth.

\section{Related Work}
\label{sec:related}
\xhdr{Agent benchmarks} Outcome-graded agent benchmarks cover software engineering, terminal work and machine-learning research \citep{jimenez2024swebench,deng2026swebenchpro,merrill2026terminalbench,chan2025mlebench,starace2025paperbench}, and GDPval and Agents' Last Exam extend evaluation to professional deliverables judged by experts \citep{patwardhan2026gdpval,sun2026ale}; checklists for agent benchmarks call for frozen inputs and validated evaluators \citep{zhu2025abc,agentsthatmatter2025}. \benchmark keeps outcome-graded tasks, but because an edit has no single correct answer, its quality test is calibrated on blind professional judgments.

\xhdr{Video editing and media agents} Video benchmarks measure components of editing, such as recognizing techniques or choosing cuts from one long video \citep{deng2026vebench,ogata2026meditbench}, or evaluate post-production operations and GUI trajectories in media software and analyze their technical failures \citep{cao2026agenticvbench,hu2026cutverse,heo2026mmtb,ai2026prosoftarena}; dedicated systems build editing agents \citep{sandovalcastaneda2025editduet,videoagent2026}. Video-understanding benchmarks ask multiple-choice questions about long or audio-visual video \citep{fu2025videomme,wu2024longvideobench,hong2026worldsense}, whereas the agents we evaluate perceive footage through still frames and transcripts. \benchmark evaluates complete assignments from raw material to a delivered edit, compares coding and GUI agents, and characterizes failures of craft from editors' notes.

\xhdr{Judging subjective outputs} Human preference ranks systems \citep{chiang2024chatbotarena,patwardhan2026gdpval}, and model judges approximate it \citep{zheng2023judging} but are biased by position \citep{wang2024fair,shi2025positionbias}, also for video \citep{ogata2026meditbench}. We therefore score each edit with a cross-laboratory panel calibrated on human judgments and validated per agent (\appref{app:related}).

\section{Conclusion}
\label{sec:conclusion}
We introduce \benchmark, a benchmark of 56 complete video-editing assignments in which agents turn raw production material and a brief into finished edits. Across 16 agents, the strongest resolves 15 tasks, while the benchmark’s quality test closely tracks human preference ($\rho=0.93$), revealing that the main gap is craft rather than compliance: 562 of 771 unresolved runs pass every other requirement but fail quality. Agents cut at just 0.72× the reference pace, rely heavily on still frames and transcripts, inspect renders for technical defects rather than pacing or story, and claim success in 93\% of runs. The results suggest that creative benchmarks need human-calibrated quality evaluation, while capable agents need native audio-visual perception, finishing tools, and the ability to judge creative quality, not merely correctness. We release the tasks, verifier, and per-run results to enable rigorous measurement of progress.

\label{sec:main-end}
\section*{Reproducibility Statement}
Sections~\ref{sec:benchmark}--\ref{sec:setup} define the tasks, tests, quality test, human study and agents. \appref{app:runtime} gives the exact model identifiers, harness versions, reasoning settings and launch commands, \appref{app:qc} the task-verification procedure, \appref{app:stats} the statistical methods, and \appref{app:release} the release. The project site (\url{https://timelinebench.tensortest.com}) and its code mirror (\url{https://timelinebench.tensortest.com/code}) provide the 56 Harbor tasks, whose briefs, tests and oracle solutions are byte-identical to those the coding agents were evaluated against; the verifier with its frozen constants and the reference edits' panel scores, so that new edits can be scored without the reference edits; every per-run outcome; and a script that recomputes the automatic-evaluation results and the count-based human-study statistics without model calls, network access, media or Docker. Statistics that need per-judgment editor records, such as the two-way editor intervals, cannot be recomputed from the release, because those records are withheld to protect participants. Judges called again may answer slightly differently, since provider serving is not guaranteed to be identical over time. Media access is gated (\appref{app:release}) and is not needed to run the recomputation script.

\section*{Ethics Statement}
The human study asked 43 freelance video editors for blind preference judgments between two complete edits. Human editors were recruited individually, and their pay did not depend on their answers. Before the first comparison, each human editor read an on-screen description of the task and of the interaction data recorded (playback controls, viewing coverage, answer changes, clicks, tab visibility and time spent; no camera, microphone, screen or keystrokes), and gave consent through a required acknowledgment. We did not seek institutional ethics review. The study collected expert preference judgments about non-sensitive video material, and apart from the contact details used to administer the study and the interaction data listed above, the only personal information collected was self-reported editing experience. We release only aggregate statistics and per-run vote counts; no editor identities, per-person telemetry or per-judgment records are published, and human editors' free-text notes appear only as short anonymous excerpts in this paper.

The EditStock project packages were purchased under license, the commissioned productions were made for the benchmark with rights to the material, and the Cinestudy projects are publicly available editing exercises. None of the source footage, reference edits or agent outputs is redistributed publicly. Identifiable people in the commissioned material are covered by releases, and the access terms forbid identifying, profiling or imitating the people shown (\appref{app:release}). Our results describe the sampled agents and tasks, not the abilities or employability of video editors.

\section*{AI Use Statement}
In this work, AI models are part of the method, and we used generative AI tools for implementation, analysis and writing. Within the method, the 16 evaluated agents are AI systems; a panel of three AI judges scores edits for the quality test and decides the judge-based brief checks (\autoref{sec:quality-test}, \appref{app:brief}); a speech-recognition model transcribes edits for code-based brief checks; the brief-test rubrics were drafted by one AI agent, adversarially reviewed by a second and verified by the authors, and are further validated by the oracle rule, single-defect controls and cross-laboratory judge agreement; two model annotators (Claude Opus 5.5 and Claude Sonnet 5) and a third adjudicating model pass tagged the editing skills of each task, and the authors cross-checked the tags (\appref{app:skills}); Claude Opus 5.5 coded the human editors' notes, with Claude Sonnet 5 as second coder, and labeled the agents' final messages, and the authors checked these codes and labels (\appref{app:analysis}); and GPT-6 Astra labeled shell-command failures, and the authors checked those labels (\appref{app:trajectory}). We also used generative AI tools to write and test code for the analyses and figures, to search and synthesize the literature, and to draft and edit this manuscript. Formulating mathematical claims and writing proofs are not applicable to this work. We have reviewed all AI-assisted work: every citation was checked against its arXiv, publisher or vendor record, two independent recomputations reproduce every deterministic number in the result files with no discrepancies, and the authors read and revised all AI-assisted text, code and analyses. We take responsibility for the final content of this work, including text, claims or artifacts produced with the aid of generative AI.

\bibliography{\subpath submission/references}
\bibliographystyle{\subpath iclr2027_conference}
\clearpage
\appendix
\startcontents[appendix]
\section*{Appendix Contents}
\printcontents[appendix]{l}{1}{\setcounter{tocdepth}{2}\setlength{\parskip}{0pt}}
\clearpage
\section{Task Composition}
\label{app:tasks}

\autoref{tab:composition} gives per-collection statistics of \benchmark.

\begin{table}[h!]
  \caption{\textbf{Composition of \benchmark by collection.} Source hours sum each task's primary moving-image footage. Medians are per task; source minutes and the source/reference ratio exclude \task{Beauty of Delhi}, which is built from 603 stills. Video files count every video container in a task's inputs. L/P: landscape/portrait; HD: 1920$\times$1080 or 1080$\times$1920; 4K: 3840$\times$2160 or 4096$\times$2160.}
  \label{tab:composition}
  \centering\footnotesize
  \setlength{\tabcolsep}{3pt}
  \begin{tabular}{@{}lrrrrrrlll@{}}
    \toprule
    & & Total & \multicolumn{4}{c}{Median per task} & \multicolumn{3}{c}{Delivery} \\
    \cmidrule(lr){3-3}\cmidrule(lr){4-7}\cmidrule(l){8-10}
    Collection & Tasks & \makecell[r]{source\\hours} & \makecell[r]{source\\minutes} & \makecell[r]{reference\\seconds} & \makecell[r]{source/\\reference} & \makecell[r]{video\\files} & format & fps & \makecell[l]{window\\(s)} \\
    \midrule
    EditStock & 11 & 18.8 & 69.5 & 67.1 & 57.3 & 96 & L; HD, 4K & 23.976, 24 & 29--305 \\
    Cinestudy & 15 & 10.3 & 27.0 & 110.9 & 14.6 & 2 & L; HD & 23.976, 24, 25 & 35--315 \\
    UGC & 15 & 1.4 & 5.7 & 39.1 & 8.4 & 50 & P; HD & 25 & 25--55 \\
    Commercial & 15 & 2.4 & 9.6 & 63.1 & 8.9 & 76 & L; HD & 24 & 44--125 \\
    \midrule
    All & 56 & 33.0 & 12.4 & 60.3 & 12.0 & 55.5 & 41 L, 15 P & 23.976, 24, 25 & 25--315 \\
    \bottomrule
  \end{tabular}
\end{table}

\subsection{Editing Skills}
\label{app:skills}

Two model annotators, Claude Opus 5.5 and Claude Sonnet 5, independently tagged each of the 56 tasks with the 16 editing skills of \autoref{fig:skills}. For each task they read the brief, the requirements of its brief-test rubric and an inventory of the supplied material, and every tag had to cite a short verbatim quote from this evidence. A skill counts as required only if the brief or its paperwork asks for it, or if the supplied material and the brief make it unavoidable; a skill that would merely be good practice is not tagged. The annotators agree on 831 of the 896 task--skill pairs (92.7\%; Cohen's $\kappa = 0.85$). A third, adjudicating model pass resolved the 65 disagreements against the quoted evidence, and the authors cross-checked every tag against its quoted evidence.

\FloatBarrier
\section{Task Verification and Quality Control}
\label{app:qc}

\begin{figure}[t]
  \centering
  \includegraphics[width=\linewidth]{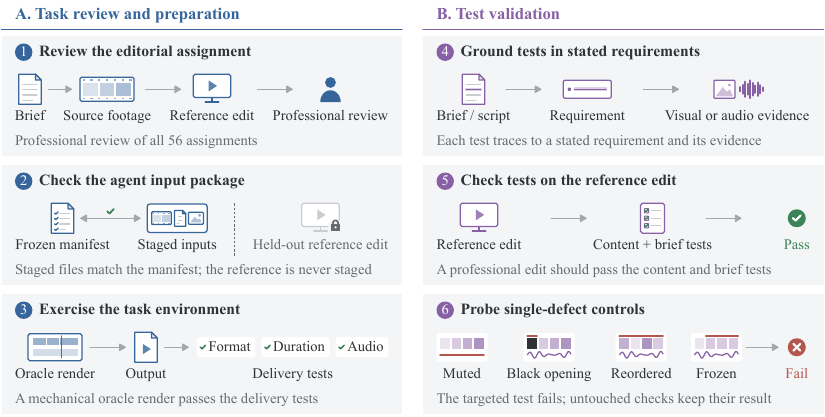}
  \caption{\textbf{Our task verification process.} (A) Each assignment is reviewed for consistency between brief, source assets and reference edit; its staged inputs are matched to a frozen manifest with the reference edit held out; and a mechanical oracle exercises the delivery tests. (B) Each content and brief test is grounded in the brief and in visual or audio evidence, checked on the reference edit, and probed with single-defect controls that should fail the targeted test.}
  \label{fig:verification}
\end{figure}

\autoref{fig:verification} summarizes the verification procedure of \autoref{sec:verification}.

\xhdr{Retrieval rule} Every brief contains two paragraphs common to all tasks, which point the agent to the tool documentation and state: ``Do not search for, retrieve, watch, or copy finished edits or reference videos for this project, including public examples. Web search and fetching are allowed for tool documentation and general editing techniques.''

\xhdr{Oracles} Each task's oracle builds a mechanical assembly: a few supplied shots, normalized and concatenated or looped to the middle of the duration window, with a supplied audio stem underneath.

\xhdr{Reference isolation} The MD5 hashes of all 56 original reference files match none of the 5,003 task input files, so no reference edit is staged in any workspace. We scanned the complete traces of all 896 runs, including tool calls and their outputs. No run made a web-fetch call, downloaded a reference edit or accessed the reference store; the only web activity is 29 web searches in 18 computer-use runs, and no reference identifier appears in their traces.

\FloatBarrier

\section{Content Tests}
\label{app:tests}

\autoref{tab:app-guards} lists the six content tests and their limits.

\begin{table}[h!]
\caption{\textbf{Content tests.} Each content test fails a video whose measurement exceeds the limit.}
\label{tab:app-guards}
\centering
\footnotesize
\begin{tabularx}{\linewidth}{@{}lYr@{}}
\toprule
Content test & Measurement & Limit \\
\midrule
Mostly silent & share of the runtime that is silent (no audio track also fails) & 0.50 \\
Mostly frozen & share of the runtime with a frozen picture & 0.60 \\
Repeated footage & duration of footage shown more than once & 1\,s \\
Dead air & longest internal silence & 5\,s \\
Channel imbalance & level difference between the left and right channels & 6\,dB \\
Faces cut by the frame edge & share of frames with a face in which a face is cut by the edge & 0.30 \\
\bottomrule
\end{tabularx}
\end{table}

\section{Quality Test}
\label{app:quality}

\subsection{Rubric}
\label{app:quality-judges}

The judges receive the study rubric verbatim.

\begin{tcolorbox}[tbbox]
\textbf{Study rubric}\\[2pt]
\emph{Purpose:} ``We are comparing complete video edits for their stated audience and purpose. There is no expected winner.''
\emph{Scope:} ``Assess editorial craft. Mandatory brand, asset, and technical delivery requirements are evaluated separately. Use the same stated audience and purpose for both versions.''
\emph{Acceptable:} ``The edit communicates its message clearly, with coherent assembly, suitable pacing, and effective picture and sound. Different creative approaches can all be acceptable.''
\emph{Needs revision:} ``An editorial weakness meaningfully harms the viewing experience: for example, confusing structure, distracting repetition, unsuitable pacing, or disruptive picture or sound. Judge viewer impact, not the time needed to fix it.''
\emph{Minor issues:} ``Optional stylistic changes and minor imperfections alone do not require rejection. `Both acceptable' does not mean equal quality. Either or both versions may need revision.''
\emph{Acceptability question:} ``Does this version meet a professional standard of editorial quality for the stated audience and purpose?''
\end{tcolorbox}

\subsection{Scoring}
\label{app:quality-scoring}

Each pass is a single call in which the judge returns integer scores from 1 to 10 for five dimensions (story and assembly, pacing, picture, sound and graphics), up to five timestamped defects, a yes-or-no answer to the acceptability question and an integer overall score from 1 to 10, where 10 is the best professional edit to expect for the project, 6 is acceptable and 3 or lower needs substantial revision.
A judge's score of an edit is $\overline{o}+0.2\,\overline{d}$, where $\overline{o}$ is the overall score and $\overline{d}$ the mean of the five dimension scores, both averaged over the judge's passes (Gemini 3.8 Flash makes two); the dimension term mainly breaks ties between equal overall scores, and the defects and the acceptability answer are not used.

\subsection{Margin Calibration}
\label{app:quality-margins}

The tie margin $t_C$ of collection $C$ is fitted on the collection's study pairs, each a delivered agent edit with its task's reference edit, and on the human editors' assessable votes on those pairs. Let $q_C$ be the share of these votes that prefer the agent edit or report no meaningful preference as per human judgement, and let the gap $g$ of a pair be the panel score of the agent edit minus that of the reference edit as per the judge model. Then $t_C$ is the smallest observed gap $x$ at which the share of the collection's pairs with $g\geq x$ does not exceed $q_C$. The frozen margins are fitted in this way on all 2,582 assessable votes on the 863 pairs.

\subsection{Individual Edits}
\label{app:quality-validation}

The quality test is valid per agent, not per edit: on single edits the panel agrees with the human editors' majority at $\kappa=0.15$, about as well as one human editor agrees with the majority of the others ($\kappa=0.14$), and the per-task pass rate correlates with the human per-task rate at only $\rho=0.32$ across the 56 tasks.

\section{Brief Compliance}
\label{app:brief}

Each judge-based brief test asks a local, binary question about visible content, never quality, that can be answered from the picture alone and is phrased so that yes means the requirement is met. Gemini 3.8 Flash watches the edit with sound, GPT-6 Astra reads contact sheets at 1\,fps, and Claude Opus 5.5 reads denser sheets of up to 3\,fps over the span the question covers. Each judge answers yes, no or cannot tell, and a two-of-three majority of yes-or-no answers decides the test.

\FloatBarrier

\section{Human Study}
\label{app:human}

\xhdr{Experience} Human editors reported their professional editing experience in four bands: less than one year (4), one to three years (27), four to seven years (8) and eight years or more (4).

\begin{figure}[tbp]
  \centering
  \includegraphics[width=\linewidth]{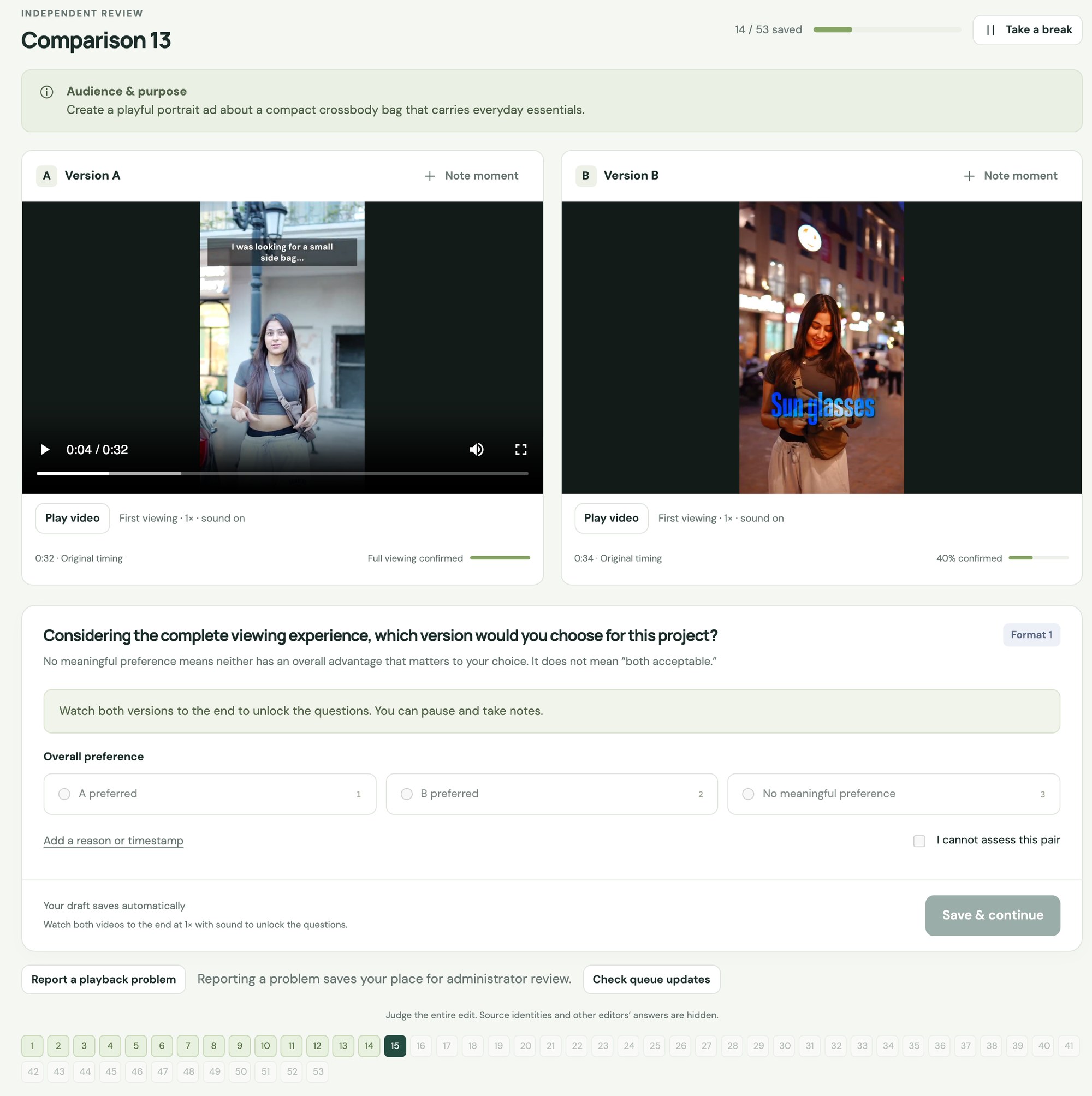}\\[4pt]
  \includegraphics[width=\linewidth]{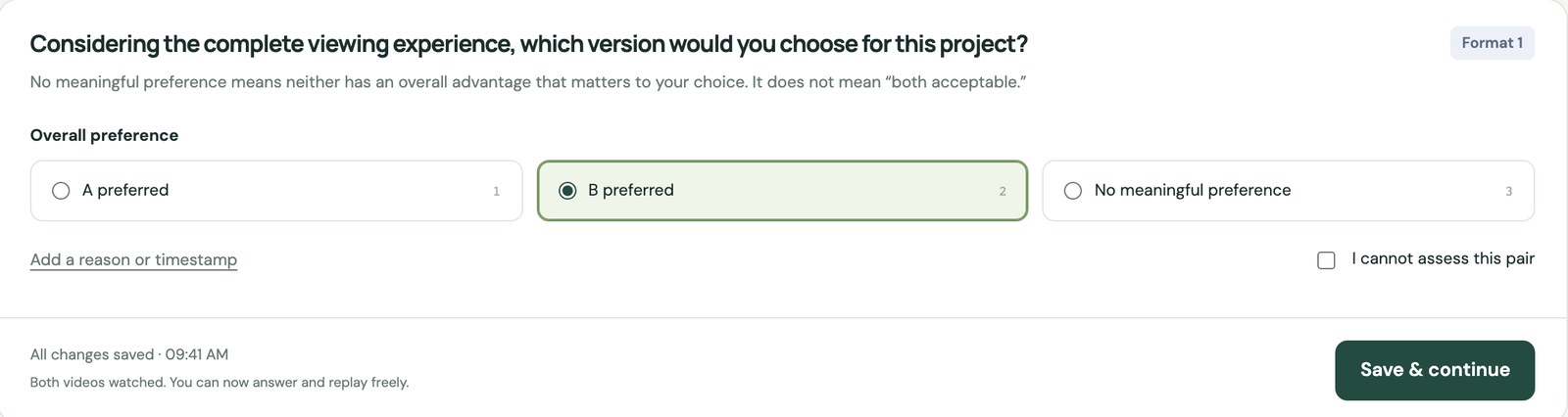}
  \caption{\textbf{The comparison page of the study interface.} Top: before both videos have been watched to the end, the answer controls are disabled; Version A has been viewed in full, so its seek bar is shown, while Version B is still in its first viewing. Bottom: the answer panel after both viewings.}
  \label{fig:study-compare}
\end{figure}

\xhdr{Order of the versions} Human editors preferred the agent edit in 10.5\% of judgments when it was Version A and in 11.7\% when it was Version B, a difference of $-1.2$ points (95\% CI $-3.7$ to $1.1$, bootstrap over tasks).

\section{Statistical Methods}
\label{app:stats}

\xhdr{Resolution rates} An agent's resolution rate is a binomial proportion over the 56 tasks, and the overall rate one over the 896 runs; both are reported with exact Clopper--Pearson 95\% intervals \citep{clopper1934}.

\xhdr{Two-way intervals} Human win-or-tie rates depend on two crossed samples, the tasks and the human editors. We therefore draw three families of 2,000 bootstrap replicates that resample tasks, human editors or individual judgments with replacement, and recompute every rate in each replicate. Let $V_{\text{task}}$, $V_{\text{editor}}$ and $V_{\text{vote}}$ be the replicate variances of a rate under the three schemes. Each one-factor bootstrap also carries the judgment-level noise, so the sum of the first two counts that noise twice \citep{owen2007pigeonhole}, and the moment-corrected estimator subtracts one copy,
\begin{equation}
\widehat{\mathrm{SE}}^2=\max\bigl(V_{\text{task}}+V_{\text{editor}}-V_{\text{vote}},\;V_{\text{task}}\bigr),
\end{equation}
where the floor keeps the interval at least as wide as the task bootstrap's. The 95\% interval is the estimate $\pm 1.96\,\widehat{\mathrm{SE}}$, truncated to $[0, 100]$.

\subsection{Reliability of Per-Agent Rates}
\label{app:reliability}

\xhdr{Single judgments} Over the 2,582 assessable judgments, Krippendorff's $\alpha$ \citep{krippendorff2018content,hayes2007krippendorff} with the ordinal metric is 0.103.

\xhdr{Per-agent rates} For every pair with three assessable judgments, we form one list of 16 win-or-tie rates from one judgment and a second list from the other two. Averaged over 3,000 random choices of the held-out judgment, the Pearson correlation of the two lists is $r = 0.6997$. Writing a rate computed from one judgment per pair as a true rate plus independent noise, $X = T + E$, with reliability $\rho_1 = \mathrm{Var}(T)/\mathrm{Var}(X)$, the two lists share only $T$, and averaging two judgments halves the noise variance, so
\begin{equation}
r = \rho_1\sqrt{\frac{2}{1+\rho_1}}, \qquad \rho_1 = \frac{r^2 + \sqrt{r^4 + 8r^2}}{4} = 0.632.
\end{equation}
The reported rates average three judgments per pair, so by the Spearman--Brown formula \citep{spearman1910,brown1910} their reliability, the correlation predicted between two lists of win-or-tie rates each computed from its own three human editors, is
\begin{equation}
\rho_3 = \frac{3\rho_1}{1+2\rho_1} = 0.838.
\end{equation}

\xhdr{Leaving out human editors} When each of the 43 human editors is left out in turn, the ranking of agents by the share of judgments preferring the agent edit keeps a Spearman correlation of $\rho \ge 0.964$ with the full ranking.

\FloatBarrier
\section{Agent Configurations and Runtime}
\label{app:runtime}

\autoref{tab:runtime-agents} gives each agent's model identifier, provider route and reasoning setting.

\begin{table}[h!]
\caption{\textbf{Agent configurations.} Model identifiers are the exact strings sent to the provider. OpenRouter routes are restricted to the named provider with fallbacks disabled. Every reasoning setting is the provider's maximum and also applies to helper and subagent roles. DeepSeek Flash is the \code{deepseek-flash} alias, served by DeepSeek-V4.1-Flash.}
\label{tab:runtime-agents}
\centering\scriptsize
\setlength{\tabcolsep}{4pt}
\begin{tabularx}{\linewidth}{@{}>{\raggedright\arraybackslash}p{2.0cm}>{\raggedright\arraybackslash}p{4.8cm}>{\raggedright\arraybackslash}p{2.1cm}Y@{}}
\toprule
Model & Model identifier & Provider route & Reasoning setting \\
\midrule
\multicolumn{4}{@{}l}{\emph{OpenCode 1.18.31}} \\
Gemini 3.1 Pro Preview & \texttt{google/gemini-3.1-pro-preview} & Google API & \texttt{thinkingLevel: high} \\
Gemini 3.8 Flash & \texttt{google/gemini-3.8-flash} & Google API & \texttt{thinkingLevel: high} \\
Claude Fable 5.1 & \texttt{anthropic/claude-fable-5-1} & Anthropic API & adaptive thinking, \texttt{effort: max} \\
Claude Opus 5 & \texttt{anthropic/claude-opus-5} & Anthropic API & adaptive thinking, \texttt{effort: max} \\
GPT-5.6 Sol & \texttt{openai/gpt-5.6-sol} & OpenAI API & \texttt{reasoningEffort: max} \\
GPT-6 Astra & \texttt{openai/gpt-6-astra} & OpenAI API & \texttt{reasoningEffort: max} \\
DeepSeek Flash & \texttt{deepseek/deepseek-flash} & DeepSeek API & \texttt{reasoning\_effort: max} \\
Grok 4.6 & \texttt{openrouter/x-ai/grok-4.6} & OpenRouter, xAI & \texttt{reasoning.effort: xhigh} \\
GLM 5.3 Flash & \texttt{openrouter/z-ai/glm-5.3-flash} & OpenRouter, Z.ai (FP8) & \texttt{reasoning.effort: max} \\
Qwen 3.8 Max & \texttt{openrouter/qwen/qwen3.8-max-0902} & OpenRouter, Alibaba & \texttt{xhigh} \\
\midrule
\multicolumn{4}{@{}l}{\emph{Codex CLI 0.155.1}} \\
GPT-5.6 Sol & \texttt{gpt-5.6-sol} & OpenAI Responses & \texttt{reasoning.effort: max} \\
GPT-6 Astra & \texttt{gpt-6-astra} & OpenAI Responses & \texttt{reasoning.effort: max} \\
GPT-6 Astra, curated guidance & \texttt{gpt-6-astra} & OpenAI Responses & \texttt{reasoning.effort: max} \\
\midrule
\multicolumn{4}{@{}l}{\emph{Claude Code 2.1.278}} \\
Claude Fable 5.1 & \texttt{claude-fable-5-1} & Anthropic Messages & adaptive thinking, \texttt{output\_config.effort: max} \\
Claude Opus 5 & \texttt{claude-opus-5} & Anthropic Messages & adaptive thinking, \texttt{output\_config.effort: max} \\
\midrule
\multicolumn{4}{@{}l}{\emph{Codex 0.155.1 with Computer Use, DaVinci Resolve on macOS}} \\
GPT-6 Astra & \texttt{gpt-6-astra} & OpenAI API & \texttt{model\_reasoning\_effort: max}, plan mode \texttt{max} \\
\bottomrule
\end{tabularx}
\end{table}

\subsection{Launch Commands}

Each coding run stages the task's inputs, writes the brief to \code{/workspace/brief.md} byte for byte as the task's \code{instruction.md}, and launches the stock CLI without interaction, with one of these commands for OpenCode, Codex CLI and Claude Code; \emph{model} and \emph{prompt} stand for the identifier in \autoref{tab:runtime-agents} and the prompt below:
\begin{tcolorbox}[tbbox]
\ttfamily\raggedright\footnotesize
opencode run --model \emph{model} --format json \emph{prompt}\par\smallskip
codex exec --model \emph{model} --json --skip-git-repo-check --dangerously-bypass-approvals-and-sandbox \emph{prompt}\par\smallskip
claude --print --model \emph{model} --effort max --output-format stream-json --verbose --dangerously-skip-permissions --settings \emph{settings} \emph{prompt}
\end{tcolorbox}
{\raggedright OpenCode reads its configuration from the \code{OPENCODE\_CONFIG\_CONTENT} environment variable. Codex CLI is configured with \code{model\_reasoning\_effort} and \code{plan\_mode\_reasoning\_effort} set to \code{max}, and the default subagent model and effort set to the selected model and \code{max}. Claude Code runs with \code{CLAUDE\_CODE\_EFFORT\_LEVEL=max}, with the subagent model and every model alias pinned to the selected model, and with \emph{settings} that allow only the selected model.\par} The prompt is a single message, identical for every coding agent:
\begin{tcolorbox}[tbbox]
\ttfamily\raggedright
Read the brief at /workspace/brief.md and complete it. Available tools and environment are documented at /opt/benchmark-tools/README.md. Follow the brief's deliverable requirements and output paths exactly.\par\medskip
\textrm{\emph{(A resource summary measured when the container starts, with advice to fit process counts, render concurrency and batch sizes to the available CPU, memory and disk.)}}
\end{tcolorbox}

The computer-use agent is launched with:
\begin{tcolorbox}[tbbox]
\ttfamily\raggedright\footnotesize
codex exec --json --ephemeral --skip-git-repo-check --enable computer\_use -m gpt-6-astra -C \emph{workspace} -c features.shell\_tool=false -c features.code\_mode=false -c sandbox\_mode="read-only" -c model\_provider="openai" -c openai\_base\_url=\emph{OpenAI API endpoint} -c forced\_login\_method="api" -c model\_reasoning\_effort="max" -c plan\_mode\_reasoning\_effort="max" -c check\_for\_update\_on\_startup=false -c approval\_policy="never" -c service\_tier="default" -c agents.default\_subagent\_model="gpt-6-astra" -c agents.default\_subagent\_reasoning\_effort="max" -
\end{tcolorbox}
Here \emph{workspace} is the task's Mac workspace, and the final dash makes Codex read its prompt from standard input.

\subsection{Execution Environment}

\autoref{tab:runtime-env} gives the coding agents' Linux environment. The delivery tests run afterwards on the output files in the task's own image (Debian 13.7, ffprobe 7.1.5, pytest 8.4.2, Python 3.12.14), without network access and with 4 CPUs and 8\,GB of memory.

\begin{table}[h!]
\caption{\textbf{Linux execution environment of the 15 coding agents.}}
\label{tab:runtime-env}
\centering\small
\begin{tabularx}{\linewidth}{@{}>{\raggedright\arraybackslash}p{2.9cm}Y@{}}
\toprule
Component & Version or content \\
\midrule
Base image & \code{node:22.22.2-bookworm-slim} (Debian 12), linux/amd64 \\
Node.js & 22.22.2 \\
Python & 3.11 virtual environment with Pillow 11.3.0, NumPy 2.2.6, SciPy 1.15.3, soundfile 0.13.1, OpenCV (headless) 4.12.0.88 and pypdf 6.0.0 \\
FFmpeg and ffprobe & 5.1.9 (Debian 12 package) \\
Remotion & 4.0.524 with its CLI and media packages; React 19.3.0 \\
HyperFrames & 0.8.40 with GSAP 3.15.0 \\
Browser & Chrome Headless Shell 149.0.7790.0 \\
Documents & Poppler utilities (Debian 12 package) \\
Transcription & \code{transcribe} command calling AssemblyAI Universal-3.5 Pro \\
\bottomrule
\end{tabularx}
\end{table}

\subsection{Cost and Runtime}
\label{app:cost}

Runtime is the CLI's wall-clock time per run. OpenCode and Claude Code record each run's model cost. Codex CLI records token counts but no dollar cost, so we estimate its cost from the tokens at base-rate prices in dollars per million tokens (uncached input, cache read, cache write, output): 10, 1, 12.5 and 50 for GPT-6 Astra, and 4, 0.4, 5 and 20 for GPT-5.6 Sol. Output tokens include reasoning tokens.

\begin{table}[h!]
\caption{\textbf{Runtime and model cost per agent,} over all 56 runs of each agent. An asterisk marks token-based cost estimates for Codex CLI.}
\label{tab:cost-runtime}
\centering\small
\begin{tabular}{@{}llrr@{}}
\toprule
 & & Median runtime & Mean cost \\
Model & Harness / guidance & per run (min) & per run (\$) \\
\midrule
GPT-6 Astra & OpenCode & 26.2 & 9.67 \\
Claude Fable 5.1 & OpenCode & 60.6 & 16.96 \\
Claude Opus 5 & OpenCode & 51.4 & 13.83 \\
GPT-5.6 Sol & OpenCode & 27.1 & 5.96 \\
Grok 4.6 & OpenCode & 25.9 & 3.46 \\
Gemini 3.8 Flash & OpenCode & 25.1 & 6.96 \\
GLM 5.3 Flash & OpenCode & 49.4 & 0.23 \\
DeepSeek Flash & OpenCode & 33.5 & 0.13 \\
Qwen 3.8 Max & OpenCode & 48.1 & 2.39 \\
Gemini 3.1 Pro Preview & OpenCode & 15.9 & 3.45 \\
\midrule
Claude Opus 5 & Claude Code & 58.5 & 20.90 \\
GPT-6 Astra & Codex CLI & 19.6 & 8.71* \\
Claude Fable 5.1 & Claude Code & 57.0 & 41.27 \\
GPT-5.6 Sol & Codex CLI & 17.9 & 5.00* \\
\midrule
GPT-6 Astra & Codex CLI, curated guidance & 21.4 & 9.76* \\
\midrule
GPT-6 Astra & Codex CLI, computer use & 52.7 & 41.61* \\
\bottomrule
\end{tabular}
\end{table}

\FloatBarrier
\section{Detailed Results}
\label{app:results}

\begin{table}[h!]
\caption{\textbf{Main results with 95\% confidence intervals}, one run per task (\autoref{tab:results}).}
\label{tab:detail-results}
\centering
\scriptsize
\setlength{\tabcolsep}{3pt}
\begin{tabular*}{\linewidth}{@{\extracolsep{\fill}}llll@{}}
\toprule
Model & Harness / guidance & Resolution rate \% [95\% CI] & Human win-or-tie \% [95\% CI] \\
\midrule
GPT-6 Astra & OpenCode & 21.4 [11.6, 34.4] & 22.3 [14.8, 29.8] \\
Claude Fable 5.1 & OpenCode & 17.9 [8.9, 30.4] & 22.9 [15.0, 30.9] \\
Claude Opus 5 & OpenCode & 17.9 [8.9, 30.4] & 19.0 [11.6, 26.4] \\
GPT-5.6 Sol & OpenCode & 17.9 [8.9, 30.4] & 17.6 [9.4, 25.8] \\
Grok 4.6 & OpenCode & 12.5 [5.2, 24.1] & 17.0 [10.1, 23.9] \\
Gemini 3.8 Flash & OpenCode & 10.7 [4.0, 21.9] & 13.1 [7.3, 18.9] \\
GLM 5.3 Flash & OpenCode & 10.7 [4.0, 21.9] & 9.5 [4.9, 14.2] \\
DeepSeek Flash & OpenCode & 7.1 [2.0, 17.3] & 13.0 [6.9, 19.2] \\
Qwen 3.8 Max & OpenCode & 3.6 [0.4, 12.3] & 4.0 [0.2, 7.8] \\
Gemini 3.1 Pro Preview & OpenCode & 1.8 [0.0, 9.6] & 9.0 [3.9, 14.0] \\
\midrule
Claude Opus 5 & Claude Code & 23.2 [13.0, 36.4] & 23.2 [15.3, 31.1] \\
GPT-6 Astra & Codex CLI & 21.4 [11.6, 34.4] & 23.8 [16.2, 31.4] \\
Claude Fable 5.1 & Claude Code & 17.9 [8.9, 30.4] & 23.5 [15.8, 31.2] \\
GPT-5.6 Sol & Codex CLI & 8.9 [3.0, 19.6] & 13.1 [7.6, 18.6] \\
\midrule
GPT-6 Astra & Codex CLI, curated guidance & 26.8 [15.8, 40.3] & 24.4 [17.1, 31.7] \\
\midrule
GPT-6 Astra & Codex CLI, computer use & 3.6 [0.4, 12.3] & 8.2 [3.0, 13.3] \\
\midrule
All agents &  & 14.0 [11.7, 16.4] & 16.5 \\
\bottomrule
\end{tabular*}
\end{table}

\subsection{Matched Contrasts}
\label{sec:contrasts}

\autoref{tab:detail-contrasts} compares pairs of agents that differ in one component, matching their edits task by task. No harness or guidance contrast is significant, while computer use significantly lowers both the win-or-tie rate and the number of resolved tasks.

\begin{table}[h!]
\caption{\textbf{Matched contrasts}, first minus second. $n$: tasks with both edits judged. $\Delta$W/T: difference in human win-or-tie rate over these tasks, in points, with 95\% CI and Holm-adjusted paired-permutation $p$. \emph{Resolved}: tasks resolved by each agent (Holm-adjusted exact McNemar $p$).}
\label{tab:detail-contrasts}
\centering
\scriptsize
\setlength{\tabcolsep}{3pt}
\begin{tabular*}{\linewidth}{@{\extracolsep{\fill}}lrlrl@{}}
\toprule
Contrast & $n$ & $\Delta$W/T [95\% CI] & $p$ & Resolved ($p$) \\
\midrule
GPT-5.6 Sol: Codex CLI $-$ OpenCode & 55 & $-$4.2 [$-$13.2, $+$4.7] & 1.00 & 5 vs 10 (0.63) \\
GPT-6 Astra: Codex CLI $-$ OpenCode & 53 & $+$0.9 [$-$7.5, $+$9.4] & 1.00 & 12 vs 12 (1.00) \\
Claude Fable 5.1: Claude Code $-$ OpenCode & 56 & $+$0.6 [$-$9.6, $+$10.8] & 1.00 & 10 vs 10 (1.00) \\
Claude Opus 5: Claude Code $-$ OpenCode & 51 & $+$5.9 [$-$2.9, $+$14.7] & 1.00 & 13 vs 10 (1.00) \\
GPT-6 Astra: curated guidance $-$ none & 56 & $+$0.6 [$-$8.4, $+$9.6] & 1.00 & 15 vs 12 (1.00) \\
GPT-6 Astra: computer use $-$ code & 53 & $-$16.4 [$-$24.0, $-$8.7] & $<$0.001 & 2 vs 12 (0.038) \\
\bottomrule
\end{tabular*}
\end{table}

\FloatBarrier

\FloatBarrier
\section{Extended Analysis}
\label{app:ext}

\subsection{Where Runs Fail}
\label{app:ext-failure}

Only 6 of the 863 delivered videos fail a delivery test, and 762 pass every brief test. The most common brief-test failure is a required spoken line that is not heard: code agents, which hear through transcription, miss 5.9\% of required lines, and the computer-use agent, which cannot hear, misses 41.5\%.

\subsection{Why Human Editors Prefer the Reference Edits}
\label{app:ext-editors}

Per-category shares of the notes are in \autoref{tab:app-notes-categories}. When human editors prefer the agent edit, they tick story and assembly more often than when they reject it (84\% against 72\% of tagged judgments) and graphics much less often (35\% against 55\%). Human editors report slates, crew talk or retakes left in 18.5\% of the computer-use agent's losses against 1.4\% for the six agents with the highest human win-or-tie rate, and repeated footage in 18.5\% against 3.6\%. An agent edit loses 6.6 points of win-or-tie per halving of its cut rate below the reference edit's (95\% CI 3.2--10.0), whereas cutting faster than the reference edit is not penalized.

\subsection{Task Difficulty}
\label{app:ext-difficulty}

\begin{figure}[t]
\centering
\includegraphics[width=\linewidth]{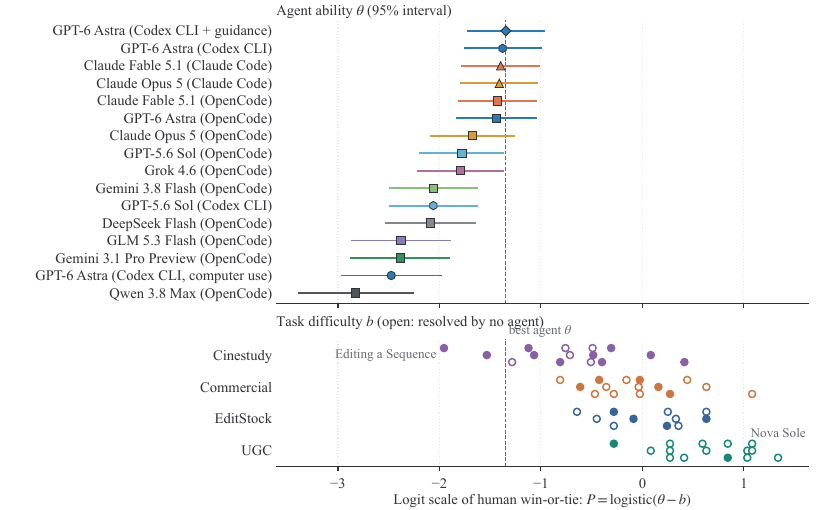}
\caption{\textbf{Tasks vary more than agents.} Agent ability (top) and task difficulty (bottom) on one logit scale, from a Rasch model of the human win-or-tie votes; an agent has even odds of a win or tie on a task whose difficulty equals its ability. Open circles: tasks resolved by no agent. The dashed line marks the best agent.}
\label{fig:wright-map}
\end{figure}

A Rasch model of the human votes, with an ability per agent and a difficulty per task (\appref{app:difficulty-model}), places agents and tasks on one scale (\autoref{fig:wright-map}). Tasks vary 2.6 times as much as agents, and on 54 of the 56 tasks even the best agent has less than even odds of a win or tie. Collection explains half of the variation in task difficulty ($\eta^2 = 0.51$). Within collections, no task descriptor or required skill predicts difficulty after Holm correction (all $|r| \le 0.24$), and once each task's difficulty is removed, the agent-by-skill grid of win-or-tie rates is at chance level (permutation $p = 0.55$).

\subsection{Trajectory-Level Analysis}
\label{app:ext-trajectory}

Within an agent, no process feature is associated with the human win-or-tie rate or with resolution after Holm correction (\autoref{fig:traj-within}). A check-then-re-render loop goes with a higher human win-or-tie rate when agents are compared on the same task ($+2.1$ points per standard deviation, Holm $p = 0.03$), but within an agent the association shrinks to $+1.5$ points and does not survive correction.

\FloatBarrier
\section{Analysis Methods}
\label{app:analysis}

\subsection{Coding of Human Editors' Notes}
\label{app:editor-notes}

\xhdr{Material} Human editors answered the optional question ``What most influenced your judgment?'' by ticking dimensions (story and assembly, pacing, picture, sound, graphics, other), by writing a note, or both. Of the 2,582 assessable judgments, 931 have at least one ticked dimension (tagged judgments), and 1,271 have a note.

\xhdr{Coding} Claude Opus 5.5 coded each note into claims, each with one of 24 categories (\autoref{tab:app-notes-categories}), a target (agent edit, reference edit or both) and a polarity (problem or strength). Comparative praise of the preferred edit is coded as a strength of that edit, not as a problem of the other.

\xhdr{Validation} Claude Sonnet 5 re-coded 150 random notes with the same prompt. On the presence of each category in a note, the two coders agree at a mean Cohen's $\kappa$ of 0.84 over the 19 categories with at least ten positives.

\begin{table}[h]
\caption{\textbf{Why human editors prefer the reference edit}, by category: share of the 1,101 notes on judgments preferring the reference edit that name a problem of the agent edit (AP), a strength of the reference edit (RS), or either (95\% interval from resampling human editors).}
\label{tab:app-notes-categories}
\centering
\scriptsize
\begin{tabular}{@{}lrrl@{}}
\toprule
Category & AP (\%) & RS (\%) & Either (\%) \\
\midrule
Overall polish & 13.8 & 46.9 & 52.8 [42.9, 63.0] \\
Shot selection & 10.0 & 29.3 & 34.2 [24.7, 42.9] \\
Text and graphics & 8.9 & 29.4 & 31.8 [24.8, 38.9] \\
Transitions and effects & 6.7 & 26.6 & 29.2 [18.4, 40.3] \\
Story and structure & 6.9 & 25.1 & 27.6 [19.8, 36.0] \\
Sound design & 5.1 & 20.9 & 23.1 [15.7, 30.6] \\
Color & 3.7 & 19.0 & 20.4 [11.7, 31.2] \\
Pacing & 2.1 & 17.2 & 18.4 [12.8, 24.4] \\
Music choice & 3.5 & 15.7 & 17.3 [11.8, 22.8] \\
Cut quality & 5.5 & 11.0 & 15.1 [10.7, 19.4] \\
Hook and opening & 2.6 & 9.7 & 10.9 \\
Framing & 2.7 & 7.7 & 9.4 \\
Branding and product & 1.6 & 8.2 & 8.6 \\
Audio defects & 7.2 & 1.6 & 8.1 [4.6, 11.9] \\
Repeated footage & 7.3 & 0.2 & 7.3 [3.0, 12.2] \\
Ending & 2.6 & 4.1 & 5.9 \\
Dialog and voice-over & 2.8 & 3.7 & 5.9 \\
Sync & 3.3 & 1.9 & 4.5 \\
Mix and levels & 1.4 & 2.9 & 4.2 \\
Text placement & 1.5 & 1.7 & 2.8 \\
Picture defects & 1.5 & 0.5 & 2.0 \\
Silence & 0.7 & 0.1 & 0.8 \\
Length & 0.1 & 0.2 & 0.3 \\
\bottomrule
\end{tabular}
\end{table}

\subsection{Task Difficulty and Skill Profiles}
\label{app:difficulty-model}

\xhdr{Rasch model} We fitted a crossed random-effects logistic model, $\mathrm{logit}\,p = \mu + \theta_{\text{agent}} - b_{\text{task}}$, to the human win-or-tie votes (863 delivered runs, three votes each) by Laplace-approximate marginal likelihood. The task standard deviation is 0.83 against 0.51 for agents, a variance ratio of 2.6.

\xhdr{Task features} We correlated the Rasch task difficulty with 20 task descriptors (collection, portrait delivery, requested duration, number and duration of source files, presence of a script, and the 16 editing skills), after removing collection means, with collection-stratified permutation tests and Holm correction.

\xhdr{Skill profiles} For each run we subtracted the mean outcome of the other 15 agents on the same task, which removes task difficulty, and compared these residuals between tasks that do and do not require each skill, within collections, testing all agent--skill cells jointly for heterogeneity by permutation.

\subsection{Trajectory Phases and Process Features}
\label{app:traj-within}

\xhdr{Phase assignment} We assigned every native action of the 896 traces to one phase by rules over the tool and its arguments. Reading the brief, documentation or file listings is orienting; frame extraction, contact sheets, image reads of source frames, \code{ffprobe} of inputs, transcription and level measurement of source files are perceiving the source; writing an edit list, plan or to-do list is planning; any command that writes the output file, or an intermediate render, is building; any command that reads, measures, transcribes or extracts frames from the agent's own render is verifying; and a build that follows a verification of the same output is repairing. Commands that match several phases take the latest phase in this order.

\begin{figure}[h]
\centering
\includegraphics[width=\linewidth]{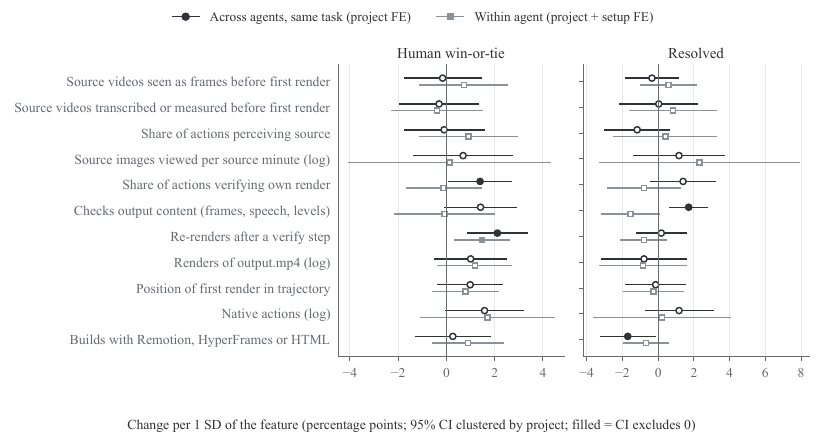}
\caption{\textbf{Process features and outcomes.} Change in the human win-or-tie rate (left) and in resolution (right) per standard deviation of each process feature, in linear probability models with task fixed effects (comparing agents on the same task; circles) and with task and agent fixed effects (comparing runs of the same agent; squares). Filled markers: the 95\% interval, clustered by task, excludes zero (before correction for multiple tests). 810 delivered code-agent runs.}
\label{fig:traj-within}
\end{figure}

\xhdr{Process and outcome} \autoref{fig:traj-within} shows eleven of the 17 process features; Holm correction is over all 34 tests (17 features, two outcomes).

\xhdr{Self-reports} Claude Opus 5.5 labeled the final agent message of each of the 896 runs as claiming full success, partial success or failure, or as a progress note without a claim, and the authors checked the labels.

\FloatBarrier
\section{Command-Failure Analysis}
\label{app:trajectory}

We labeled all 55,628 recorded shell commands of the 15 code agents with GPT-6 Astra as a judge. Of the 52,737 commands with a decided label, 6.0\% fail.

\FloatBarrier
\section{Release and Licensing}
\label{app:release}
Code, task definitions, rubrics, frozen constants and results of \benchmark 1.0 are released under Apache-2.0. Each Harbor task holds its configuration, the brief, an environment definition, an input-staging script, an oracle solution and the four delivery tests; the verifier package adds the content tests, the brief tests with their 56 rubrics, and the quality test. The results include every per-run outcome with sanitized judge outputs, and the human study is released as aggregate statistics and per-run vote counts.

Source media are available only to approved research teams, for research and evaluation. The UGC and Commercial productions were commissioned for the benchmark, with rights to the material and releases for the people, locations and brands shown. The access terms forbid redistributing inputs, outputs, stills, frames or clips; training generative models on the inputs or anything derived from them; training or tuning on the benchmark; and identifying, profiling or imitating the people shown.

\section{Related Evaluation Settings}
\label{app:related}
\autoref{tab:related-settings} compares \benchmark with the closest evaluation settings by what each receives, what it scores and how it scores it.

\begin{table}[h!]
\caption{\textbf{Related evaluation settings.}}
\label{tab:related-settings}
\centering\footnotesize
\setlength{\tabcolsep}{3pt}
\begin{tabularx}{\linewidth}{@{}>{\raggedright\arraybackslash}p{2.0cm}YYYY@{}}
\toprule
Benchmark & Input & Output scored & Evaluation & Human judgment of outputs \\
\midrule
VEBench\newline\citep{deng2026vebench} & Edited videos and a question & Answer, clip choice or time span & Accuracy; temporal overlap & None \\
MEDit-Bench\newline\citep{ogata2026meditbench} & One long video and an editing message & Cut list & Temporal overlap with professional edits & User study on a subset (1,620 evaluations) \\
AgenticVBench\newline\citep{cao2026agenticvbench} & Source videos and a brief or storyboard & Video with a manifest or report & Programmatic verifiers; 1,069 binary rubric items for repurposing & Expert rubric grading; three-editor human baseline on a subset \\
CutVerse\newline\citep{hu2026cutverse} & GUI application state and an objective & GUI trajectory & Milestone checks & Not reported \\
ProSoftArena\newline\citep{ai2026prosoftarena} & Real desktop and a task & Final state or artifact & Execution scripts; subjective comparison with human work on creative tasks & No rating reported \\
MultiMedia-\allowbreak TerminalBench\newline\citep{heo2026mmtb} & Terminal workspace with media files & File artifact & Task verifiers with binary and partial success & None \\
GDPval\newline\citep{patwardhan2026gdpval} & Request and reference files & Work product & Blinded expert pairwise comparison with expert deliverables & Occupation experts \\
\benchmark\newline(ours) & Raw production material and brief & Rendered video & Delivery tests, content tests, brief tests and a quality test calibrated on human editors, combined into a resolution rate & 43 video editors; 2,589 blind judgments against the reference edit \\
\bottomrule
\end{tabularx}
\end{table}

\end{document}